\documentclass{article}

 \usepackage[preprint]{neurips_2026}

\usepackage[utf8]{inputenc} %
\usepackage[T1]{fontenc}    %
\usepackage{hyperref}       %
\usepackage{url}            %
\usepackage{booktabs}       %
\usepackage{amsfonts}       %
\usepackage{amsmath}        %
\usepackage{nicefrac}       %
\usepackage{microtype}      %
\usepackage{xcolor}         %
\usepackage[inline]{enumitem}
\usepackage{algorithm}
\usepackage{algpseudocode}
\usepackage{graphicx}
\usepackage{subcaption}  %

\newcommand{\secsq}{\vspace{-8pt}}    %
\newcommand{\subsq}{\vspace{-5pt}}    %
\newcommand{\parsq}{\vspace{-4pt}}    %
\newcommand{\figsq}{\vspace{-8pt}}    %

\makeatletter
\newcommand{\dirharp@gap}{0ex}               %
\newcommand{\dirharp@scale}{0.8}             %
\newcommand{\dirharp}[2]{\mathpalette{\dirharp@{#1}}{#2}}
\newcommand{\dirharp@}[3]{%
  \sbox\z@{$\m@th#2#3$}%
  \ooalign{%
    \hfil$\m@th#2#3$\hfil\cr
    \hfil\raisebox{\dimexpr\ht\z@+\dirharp@gap\relax}{%
      \scalebox{\dirharp@scale}{$\m@th\scriptstyle#1$}}\hfil\cr}}
\makeatother
\newcommand{\fwdh}[1]{\dirharp{\rightharpoonup}{#1}}   %
\newcommand{\bwdh}[1]{\dirharp{\leftharpoonup}{#1}}    %

\DeclareMathOperator{\E}{\mathbb{E}}
\DeclareMathOperator{\ESS}{ESS}
\newcommand{\R}{\mathbb{R}}
\newcommand{\norm}[1]{\lVert #1 \rVert}
\newcommand{\dd}{\,d}

\definecolor{metalicacol}{HTML}{2B59C8}   %
\definecolor{metadiffcol}{HTML}{8C4A11}   %
\newcommand{\metalica}{METALICA}
\newcommand{\metadiff}{MetaDiff}
\newcommand{\metalicac}{\textcolor{metalicacol}{\textbf{\metalica}}}   %
\newcommand{\metadiffc}{\textcolor{metadiffcol}{\textbf{\metadiff}}}
\newcommand{\crepe}{CREPE}

\newcommand{\pot}{U}                       %
\newcommand{\cv}{\xi}                      %
\newcommand{\cvvar}{\omega}                %
\newcommand{\cvmarg}{p^{\cv}}              %
\newcommand{\fe}{F^{\cv}}                  %
\newcommand{\dfe}{\Delta F^{\cv}}          %
\newcommand{\feh}{\hat{F}^{\cv}}           %
\newcommand{\dfeh}{\Delta\hat{F}^{\cv}}    %
\newcommand{\region}{\Omega}               %
\newcommand{\basinA}{\Omega_{\mathrm{A}}}  %
\newcommand{\basinB}{\Omega_{\mathrm{B}}}  %

\newcommand{\pit}[1]{p_{#1}}               %
\newcommand{\score}{\nabla_x \log \pit{t}}  %
\newcommand{\denoise}{\hat{x}_{0|t}}       %
\newcommand{\denoiseat}[1]{\hat{x}_{0|#1}}  %
\newcommand{\fwd}[2]{\fwdh{F}_{#1|#2}}     %
\newcommand{\bwd}[2]{\bwdh{B}_{#1|#2}}     %

\newcommand{\tpit}[1]{\pi_{#1}}            %
\newcommand{\tbias}[1]{r_{#1}}             %
\newcommand{\pik}[2]{\pi_{#1,#2}}          %
\newcommand{\pikprod}[1]{\pik{t_0}{#1}\times\cdots\times\pik{t_M}{#1}}
\newcommand{\incw}[2]{g_{#1|#2}}           %
\newcommand{\nwt}{W}                       %

\newcommand{\bias}{V}                      %
\newcommand{\hill}{b}                      %
\newcommand{\anneal}{\lambda}              %
\newcommand{\hillh}{h}                     %
\newcommand{\hillw}{\sigma_{\mathrm{hill}}}  %
\newcommand{\biasfac}{\gamma}              %
\newcommand{\gexp}{\rho}                   %

\newcommand{\schedule}{0 = t_0 < t_1 < \cdots < t_M = 1}  %
\newcommand{\state}[1]{X^{0:M}_{#1}}      %
\newcommand{\rep}[2]{X^{#1}_{#2}}          %
\newcommand{\acc}{\alpha^{\mathrm{swap}}}  %
\newcommand{\wt}{w}                        %
\newcommand{\samples}[1]{\mathcal{X}_{#1}} %
\newcommand{\allsamples}{\mathcal{X}}      %

\newcommand{\batch}{N}                     %

\title{METALICA:\\METAdynamics and repLICA exchange for enhanced diffusion sampling}

\author{%
  Alireza Omidi$^*$ \\
  University of British Columbia \\
  \texttt{aomidi@student.ubc.ca} \\
  \And
  Jiajun He \\
  University of Cambridge \\
  \texttt{jh2383@cam.ac.uk} \\
  \AND
  Jörg Gsponer$^\dagger$ \\
  University of British Columbia \\
  \texttt{gsponer@msl.ubc.ca}
  \And
  Saifuddin Syed$^\dagger$ \\
  University of British Columbia \\
  \texttt{saif.syed@stat.ubc.ca}
}

\begin{document}
\begingroup
\renewcommand{\thefootnote}{}
\footnotetext{$^*$ Corresponding author.}
\footnotetext{$^\dagger$ Joint senior authors.}
\endgroup

\maketitle

\begin{abstract}
Many proteins function through transitions between conformational states, yet rare states are rarely sampled by diffusion models trained on an equilibrium ensemble, demanding better sampling methods. We introduce \metalica{}, which implements Metadynamics on a pretrained diffusion model via Replica Exchange. It accumulates a bias potential along a Collective Variable, repels new samples from previous ones through biased sampling, and reweights samples onto the unbiased distribution. \metalica{} holds one replica per diffusion level, forming a Markov Chain that evolves through inter-replica communication and is refined in place as the bias grows. \metalica{} is the dual of sequential control, in which Sequential Monte Carlo parallelizes the sampler over a batch of particles. Parallelism over the levels of the diffusion-time schedule instead allows \metalica{} to generate samples from long chains, essential for the discovery of rare events, with accuracy set by run length rather than by the memory available. We validate on a bimodal target with known free energies, then apply \metalica{} to the unfolding of a protein. At a budget for which sequential control yields no unfolded structure, \metalica{} populates the basin and resolves a second free energy minimum.
\end{abstract}

\secsq
\section{Introduction}
\label{sec:intro}
Proteins undergo functionally important transitions between distinct conformational states \citep{henzler-wildmanDynamicPersonalitiesProteins2007,boehrRoleDynamicConformational2009}, and are thus better represented by conformational ensembles and energy landscapes than by a single static structure \citep{onuchicTheoryProteinFolding1997}. Molecular Dynamics (MD) simulations are widely used to generate such ensembles \citep{hollingsworthMolecularDynamicsSimulation2018}, but often struggle to explore complex energy landscapes \citep{heninEnhancedSamplingMethods2022}. Enhanced sampling techniques such as Umbrella Sampling \citep{torrieNonphysicalSamplingDistributions1977} and Metadynamics \citep{laioEscapingFreeenergyMinima2002} address this by biasing along predefined Collective Variables (CVs), encouraging sampling of under-populated regions \citep{heninEnhancedSamplingMethods2022}. However, such simulations remain computationally demanding, and since samples derive from time-evolving trajectories, they are inherently correlated.

In parallel, diffusion models, often trained or fine-tuned on MD data to learn conformational distributions, have shown strong success generating uncorrelated biomolecular ensembles, including ensembles for proteins \citep{lewisScalableEmulationProtein2025}. Unlike MD, they need not overcome free-energy barriers, but since they sample from the learned equilibrium distribution, rare, yet functionally important, conformations remain hard to capture without prohibitively large sample sizes \citep{noeBoltzmannGeneratorsSampling2019}. Inspired by the enhanced sampling approaches for MD simulations, recent works have tried to solve the rare event sampling problem by integrating Umbrella Sampling \citep{richmanUnlockingHiddenBiomolecular2026} and Metadynamics \citep{xieEnhancedDiffusionSampling2026,lamMetadiffusionInferencetimeMetaenergy2026} into diffusion models. Rather than sample the learned data distribution $\pit{0}(x)$, these methods tilt the target to $\pit{0}(x)\exp(\tbias{0}(x))$, with $\tbias{0}$ being a reward function intended to boost the probability of rare states. Sampling from the tilted targets is achieved via Sequential Monte Carlo (SMC)-based methods such as the Twisted Diffusion Sampler (TDS, \citet{wuPracticalAsymptoticallyExact2023}) or the Feynman-Kac Corrector (FKC, \citet{skretaFeynmanKacCorrectorsDiffusion2025a}), where a fresh batch of samples is initialized with pure noise and steered towards each of the target tilted distributions. After gathering batches of samples, each under a different bias potential, they are pooled using the Multistate Bennett Acceptance Ratio (MBAR) \citep{shirtsStatisticallyOptimalAnalysis2008a} or Weighted Histogram Analysis Method  (WHAM) \citep{https://doi.org/10.1002/jcc.540130812} to reconstruct the unbiased distribution $\pit{0}$.

The accuracy of enhanced diffusion sampling depends on SMC-based approaches correctly sampling tilted target distributions, which requires large batch sizes and small denoising steps \citep{phillipsParticleDenoisingDiffusion2024,renDRIFTLITELIGHTWEIGHTDRIFT2026}. Otherwise, lineage collapse and reduced Effective Sample Size (ESS) degrade performance. This worsens for rare events requiring stronger tilts \citep{xieEnhancedDiffusionSampling2026}, since only a tiny fraction of samples fall in the rare region, making current implementations inefficient and inaccurate under realistic memory and compute constraints.
In response, we introduce \metalica{}, an enhanced sampling method integrating METAdynamics into diffusion models via repLICA exchange. Rather than batch sampling, \metalica{} runs a single Markov Chain with replicas at different denoising steps in parallel, sampling sequentially. 

\secsq
\section{Background}
\label{sec:background}

\subsq
\subsection{Free energy differences and Metadynamics}
\label{sec:bg_metad}

\parsq
\paragraph{Collective Variables and free energy.}
A molecular conformation is a point $x \in \R^D$ drawn from the Boltzmann density
$\pit{0}(x)$. Enhanced sampling biases along a Collective Variable (CV), a map
$\cv:\R^D \to \R^d$ with $d \ll D$ chosen to separate metastable states; we write
$\cvvar = \cv(x)$ for the CV state of a conformation, $\region \subseteq \R^d$
for a region of CV space, and mark densities on CV space with a superscript $\cv$.
The marginal of $\pit{0}$ in CV space and the free energy of a region $\region$ are
\begin{equation}\label{eq:pmf}
\cvmarg(\cvvar) \;=\; \int \pit{0}(x)\,\delta\bigl(\cvvar - \cv(x)\bigr)\dd x,
\qquad
\fe(\region) \;=\; -\log \cvmarg(\region)
\;=\; -\log \int_{\region} \cvmarg(\cvvar)\dd\cvvar ,
\end{equation}
the negative log probability that an equilibrium sample lands in $\region$; the
free energy surface $\fe(\cvvar) = -\log\cvmarg(\cvvar)$ is its density version, defined
up to an additive constant; the argument's case says which is meant, a capital
$\region$ the free energy of a region and a lower-case $\cvvar$ the surface. We
wish to estimate the free energy surface and the free energy difference
$\dfe = \fe(\basinB) - \fe(\basinA) = \log \cvmarg(\basinA)/\cvmarg(\basinB)$
between two regions $\basinA$ and $\basinB$ of CV space, a populated and a rare
metastable state.

\parsq
\paragraph{Metadynamics.}
Metadynamics \citep{laioEscapingFreeenergyMinima2002} escapes free energy minima by
accumulating a bias potential on CV space that penalizes regions already sampled.
Let $\bias_k:\R^d \to \R$ be the bias used in iteration $k = 0,\dots,K$, with
$\bias_0 \equiv 0$. Iteration $k$ draws conformations from the biased density and
adds a non-negative increment $\hill_k:\R^d \to \R$ concentrated where they landed,
\begin{equation}\label{eq:metad}
\pik{0}{k}(x) \;\propto\; \pit{0}(x)\,\exp\bigl(-\bias_{k}(\cv(x))\bigr),
\qquad
\bias_{k+1} \;=\; \bias_{k} + \hill_k .
\end{equation}
The increments are chosen so that $\bias_k$ converges while successive increments
stay small; a common choice is Well-Tempered Metadynamics (WTMetaD)
\citep{barducciWellTemperedMetadynamicsSmoothly2008}, a Gaussian hill per sample
whose height decays where the bias has grown (Appendix~\ref{app:wtmetad}). Rather
than reading $\fe$ off the converged bias, we pool the samples drawn under
$\pik{0}{0},\dots,\pik{0}{K}$ and reweight them onto $\pit{0}$ with the multistate
Bennett acceptance ratio (MBAR, Appendix~\ref{app:mbar}).

\subsq
\subsection{Inference-time control of diffusion models}
\label{sec:bg_control}

A diffusion model \citep{songSCOREBASEDGENERATIVEMODELING2021} defines a path of
marginals $\{\pit{t}\}_{t\in[0,1]}$ from the data distribution $\pit{0}$ to a tractable
noise distribution $\pit{1}$ as the law of a forward SDE, together with a score
$\score$ that we assume available; samples of $\pit{0}$ are generated by integrating
the reverse SDE from $\pit{1}$,
\begin{equation}\label{eq:sdes}
\dd X_t = f_t(X_t)\dd t + \sigma_t\dd\fwdh{W}_t,
\qquad
\dd X_t = \bigl[f_t(X_t) - \sigma_t^2\,\nabla_x\log\pit{t}(X_t)\bigr]\dd t
 + \sigma_t\dd\bwdh{W}_t,
\end{equation}
with $\fwdh{W}$ and $\bwdh{W}$ Brownian motions forward and backward in time.
\emph{Inference-time control} aims to sample from a tilted target
$\pit{0}\exp(\tbias{0})$, for a reward $\tbias{0}$ on $\R^D$, without retraining. It
extends the reward $\tbias{t}$ to diffusion time and targets the tilted path
\begin{equation}\label{eq:tilted_path}
\tpit{t}(x) \;\propto\; \pit{t}(x)\exp\bigl(\tbias{t}(x)\bigr),
\qquad
\tbias{t}(x) \;=\; \anneal_t\,\tbias{0}\bigl(\denoise(x)\bigr),
\end{equation}
where $\denoise(x) = \E[x_0 \mid x_t = x]$ is the denoiser, available from the score
by Tweedie's formula, with $\denoiseat{0}(x) = x$, and $\anneal_t$ is a guidance
schedule decreasing from $\anneal_0 = 1$ to $\anneal_1 = 0$ (we use
$\anneal_t = (1-t)^{\gexp}$, Appendix~\ref{app:method}): the path starts at the
noise distribution, $\tpit{1} = \pit{1}$, and ends at the target. Its score
$\nabla_x\log\tpit{t} = \nabla_x\log\pit{t} + \nabla_x\tbias{t}$ is available, so the
reverse SDE \eqref{eq:sdes} can be run with $\tpit{t}$ in place of $\pit{t}$, but the
resulting marginals from this \emph{guided} SDE are not $\tpit{t}$ and a correction is needed.

One commonly used strategy is to correct the move with a weight.
Fix a schedule $\schedule$ and let $\fwd{m}{m-1}$ and $\bwd{m-1}{m}$ be forward and
backward proposal kernels between levels $m-1$ and $m$ (the two SDEs of
\eqref{eq:sdes} integrated between $t_{m-1}$ and $t_m$). The
\emph{incremental weight} of a backward move $x_m \to x_{m-1}$ is
\begin{equation}\label{eq:incw}
\incw{m-1}{m}(x_{m-1}, x_m) \;=\;
\frac{\tpit{t_{m-1}}(x_{m-1})\,\fwd{m}{m-1}(x_m \mid x_{m-1})}
     {\tpit{t_m}(x_m)\,\bwd{m-1}{m}(x_{m-1} \mid x_m)},
\end{equation}
the density the tilted path assigns to the pair over the density the proposals
assign to it; the reward enters as $\exp(\tbias{t_{m-1}}(x_{m-1}) - \tbias{t_m}(x_m))$
and the rest is computable from the score along the move
\citep{heRNEPlugandplayDiffusion2025} (Appendix~\ref{app:crepe}). The error of each
backward move can be corrected with \eqref{eq:incw} either sequentially or in parallel.

\parsq
\paragraph{Sequential control.}
Sequential Monte Carlo samplers (SMC) \citep{delmoralSequentialMonteCarlo2006}
applied to \eqref{eq:incw} give the sequential control methods, like FKC, TDS, RNE and
DriftLite
\citep{skretaFeynmanKacCorrectorsDiffusion2025a,wuPracticalAsymptoticallyExact2023,heRNEPlugandplayDiffusion2025,renDRIFTLITELIGHTWEIGHTDRIFT2026}.
A batch of $\batch$ particles $\rep{1:\batch}{M} \sim \pit{1}$ is drawn from noise
with uniform weights $\wt^{1:\batch}_M = 1$; throughout, a superscript indexes what the
sampler runs in parallel and a subscript what it advances in sequence, so particles are
superscripts here and levels are superscripts in \eqref{eq:pt_proposal}. Step $m$ propagates the
particles with the backward kernel and the weights with the incremental weight,
\begin{equation}\label{eq:smc_step}
\rep{n}{m-1} \;\sim\; \bwd{m-1}{m}(\cdot \mid \rep{n}{m}),
\qquad
\wt^{n}_{m-1} \;=\; \wt^{n}_m\;\incw{m-1}{m}(\rep{n}{m-1}, \rep{n}{m}),
\end{equation}
optionally resampling $\rep{1:\batch}{m-1}$ from the normalized weights
$\nwt^{1:\batch}_{m-1}$; we write $\wt$ for weights and $\nwt$ for normalized weights
throughout. At $m = 0$ the weighted batch targets $\tpit{0}$, and self-normalized
estimates under $\tpit{0}$ are biased for finite $\batch$ but consistent as
$\batch \to \infty$.

\parsq
\paragraph{Parallel control.}
Accelerated Parallel Tempering (APT)
\citep{ballardReplicaExchangeNonequilibrium2009,zhangAcceleratedParallelTempering2026}
applied to \eqref{eq:incw} gives \crepe{} \citep{heCREPECONTROLLINGDIFFUSION2026},
which keeps one particle per level and runs a Replica Exchange (RE) chain
\citep{swendsenReplicaMonteCarlo1986,geyerMarkovChainMonte1991,hukushimaExchangeMonteCarlo1996}
$\state{n} = (\rep{0}{n},\dots,\rep{M}{n})$ targeting
$\tpit{t_0}\times\cdots\times\tpit{t_M}$, with the schedule over diffusion time
rather than temperature. Given $(\rep{m-1}{n}, \rep{m}{n})$ at adjacent levels, the
forward proposal moves the lower particle up and the backward proposal moves the
upper particle down,
\begin{equation}\label{eq:pt_proposal}
X'^{\,m} \;\sim\; \fwd{m}{m-1}(\cdot\mid \rep{m-1}{n}),
\qquad
X'^{\,m-1} \;\sim\; \bwd{m-1}{m}(\cdot\mid \rep{m}{n}),
\end{equation}
and the pair is exchanged, $(\rep{m-1}{n+1},\rep{m}{n+1}) = (X'^{\,m-1}, X'^{\,m})$,
with probability
\begin{equation}\label{eq:swap}
\acc_{m-1,m} \;=\; \min\left\{1,\;
\frac{\incw{m-1}{m}(X'^{\,m-1}, \rep{m}{n})}{\incw{m-1}{m}(\rep{m-1}{n}, X'^{\,m})}\right\},
\end{equation}
the incremental weight of the proposed backward move over that of the move it
reverses; otherwise nothing happens and
$(\rep{m-1}{n+1},\rep{m}{n+1}) = (\rep{m-1}{n},\rep{m}{n})$. At step $n$ the pairs
with $m \equiv n \pmod 2$ are proposed in parallel, the non-reversible scheme of
\citet{syedNonReversibleParallelTempering2022}; the reference particle $\rep{M}{n}$
is redrawn from $\pit{1}$ and the target particle $\rep{0}{n}$ is recorded as a
sample of $\tpit{0}$. Running $\batch$ steps yields $\batch$ correlated samples, and
averages over them are biased by the burn-in but consistent as $\batch \to \infty$.

\secsq
\section{Method}
\label{sec:method}

Metadynamics on a diffusion model alternates two steps: sample the tilted target
\eqref{eq:metad} of the current bias by inference-time control, then deposit an
increment from those samples. \metadiff{}
\citep{xieEnhancedDiffusionSampling2026} samples with sequential control, a fresh
batch of $\batch$ particles annealed from $\pit{1}$ with FKC at every iteration.
\metalica{} samples with parallel control, a single \crepe{} chain carried across all
iterations whose target changes at every bias update. Algorithm~\ref{alg:metalica}
states both; they differ in one line.

At iteration $k$ the bias is $\bias_k$ and the tilted path is \eqref{eq:tilted_path}
with reward $\tbias{0} = -\bias_k \circ \cv$,
\begin{equation}\label{eq:intermediate_bias}
\pik{t}{k}(x) \;\propto\; \pit{t}(x)\exp\Bigl(-\anneal_t\, \bias_{k}\bigl(\cv(\denoise(x))\bigr)\Bigr),
\end{equation}
where the subscript $t$ is diffusion time and $k$ the iteration; since
$\bias_0 \equiv 0$, $\pik{t}{0} = \pit{t}$ and the first iteration is unbiased.
\metalica{} advances the RE chain $\state{n}$ for $\batch$ steps against
$\pikprod{k}$ on a schedule uniform in $t$, with \eqref{eq:swap} evaluated for
$\pik{t}{k}$; the denoiser at a proposal is a by-product of the score evaluation that
generated it, so an exchange costs one backward transition
(Appendix~\ref{app:crepe}). The target particles visited during iteration $k$,
$\samples{k} = \{\rep{0}{k\batch+n}\}_{n=1}^{\batch}$, are $\batch$ consecutive states of one chain,
so the $K+1$ iterations are segments of a single run rather than $K+1$ runs. The
increment $\hill_k$ is deposited from $\samples{k}$ by any Metadynamics rule and
$\bias_{k+1} = \bias_k + \hill_k$; the experiments use the Well-Tempered rule
(Appendix~\ref{app:wtmetad}). Finally the pooled samples
$\allsamples = \bigcup_k \samples{k}$, drawn from $\pik{0}{0},\dots,\pik{0}{K}$ with
known biases $\bias_0,\dots,\bias_{K}$, are reweighted onto $\pit{0}$ by MBAR
(Appendix~\ref{app:mbar}).

\begin{algorithm}[tbp]
\caption{Metadynamics with inference-time control: \metadiffc{} and \metalicac{}}
\label{alg:metalica}
\begin{algorithmic}[1]
\Require Score $\score$; CV $\cv$; bias-update rule; iterations $K$; $\batch$ (SMC
particles or RE steps per iteration); schedule $\schedule$.
\State $\bias_0 \gets 0$; $\allsamples \gets \emptyset$; \metalicac{} only:
initialize the RE chain $\state{0}$ by one reverse pass from $\pit{1}$
\For{$k = 0, \ldots, K$}
    \State Set the target path $\pik{t}{k}$ by \eqref{eq:intermediate_bias} using $\bias_k$
    \State \metadiffc{}: run SMC \eqref{eq:smc_step} with $\batch$ fresh particles from
    $\pit{1}$; $\samples{k} \gets$ the weighted batch at $m = 0$
    \State \metalicac{}: advance the RE chain $\batch$ steps with \eqref{eq:pt_proposal},
    \eqref{eq:swap}; $\samples{k} \gets$ the $\batch$ visited target particles
    \Comment{chain not reset}
    \State $\allsamples \gets \allsamples \cup \samples{k}$;\quad
    deposit $\hill_k$ from $\samples{k}$;\quad $\bias_{k+1} \gets \bias_{k} + \hill_k$
\EndFor
\State \Return $\allsamples$ and weights $\wt \gets \textsc{MBAR}\bigl(\allsamples, \{\bias_{k}\}_{k=0}^{K}\bigr)$
\end{algorithmic}
\end{algorithm}

\parsq
\paragraph{Online refinement.}
Since the exponent of \eqref{eq:intermediate_bias} is linear in $\bias_k$,
consecutive paths differ by one increment (Appendix~\ref{app:recursion}): for
$k = 0,\dots,K-1$,
\begin{equation}\label{eq:recursion}
\pik{t}{k+1}(x) \;\propto\; \pik{t}{k}(x)\,
\exp\Bigl(-\anneal_t\, \hill_{k}\bigl(\cv(\denoise(x))\bigr)\Bigr).
\end{equation}
The increment is small by construction, since tempering shrinks
$\hill_{k}$ as the bias fills in. Because an RE chain is just a state, its target can
be changed mid-run, and the current state remains a well-equilibrated start for the
new target when the change is small. \metalica{} therefore refines the target in
place: the chain carries over from iteration $k$ to $k+1$ with no fresh burn-in,
which sequential control cannot do since it restarts from noise at every $k$.

\parsq
\paragraph{A computational duality.}
\metadiff{} and \metalica{} accept any bias-update rule and share \eqref{eq:metad};
they differ only in which axis the sampler parallelizes over, and that decides what
one pays for more samples. Under sequential control, more samples means more
particles: memory grows with $\batch$, wall-clock does not, and the samples are nearly
independent, but every particle is annealed from noise again at each bias update,
so accuracy at each iteration is whatever a batch of size $\batch$ can deliver. Under
parallel control, more samples means more steps of the same chain: wall-clock grows
with $\batch$, memory does not, and each step starts from an equilibrated state and pays
no restart at a bias update \eqref{eq:recursion}, so accuracy is set by chain length
rather than by memory. The two spend the same $\batch M$ score evaluations per iteration
and neither dominates: \metadiff{} is the right tool when memory is plentiful and
time is short, \metalica{} when memory is the binding constraint, which is the case
for rare events, where a batch must be large enough to represent the rare region
before any bias can be deposited from it. Appendix~\ref{app:toy_experiment} measures this trade-off
(Figures~\ref{fig:toy_bs} and~\ref{fig:toy_memory_time}).

\secsq
\section{Experiments}\label{sec:experiments}

\parsq
\paragraph{1D Toy.}
We first test \metalica{} in a toy setting where $\pit{0}$ is known: a bimodal Gaussian mixture on $\R$ with modes centered at $\mu = \pm 3$ of equal width ($\sigma_{\mathrm{mix}} = 0.5$). We take the CV to be the data coordinate itself, $\cv(x) = x$, so that $\basinA = \{\cvvar < 0\}$ holds the populated mode and $\basinB = \{\cvvar > 0\}$ the rare one. The free energy difference between the two modes is $\dfe = \eta \log 10$, so the rare mode is exactly $10^{\eta}$ times less populated than the populated mode.

\begin{figure}[tbp]
    \centering
    \includegraphics[width=\linewidth]{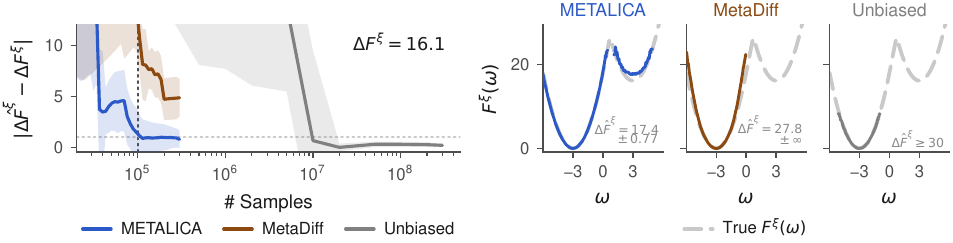}
    \caption{Free energy estimation on the 1D toy. \textbf{Left:} error in $\dfeh$ against cumulative samples. \textbf{Right:} the surface $\feh$ each method reconstructs at $100{,}000$ samples, against the true $\fe$ (dashed).}
    \label{fig:toy}\figsq
\end{figure}

To simulate the rare event problem, we set $\eta = 7$ $\bigl(\dfe \approx 16.1\bigr)$, which results in a highly unbalanced free energy landscape where the rare mode is ten million times less probable than the likelier one. We then set the bias factor $\biasfac = 10$, iteration sample size $\batch = 100$, and draw $100{,}000$ samples through $K = 1{,}000$ iterations. To compare, we also run \metadiff{} with the same parameters, and unbiased sampling, i.e.\ the same diffusion model run with no bias and no steering. Throughout the iterations, we pool the cumulative samples and unbias them using MBAR, and calculate the intermediate error $|\dfeh - \dfe|$ to track convergence, where $\dfeh$ is the free energy difference of Section~\ref{sec:bg_metad} evaluated on the MBAR-reconstructed profile $\feh$ (Figure~\ref{fig:toy}).

To compare how iteration sample size affects the performance of both \metalica{} and \metadiff{}, we try larger $\batch$ values up to $20{,}000$ while keeping the total sampling budget fixed at $100{,}000$, and follow reconstruction quality, memory and computation time (Figures~\ref{fig:toy_bs} and~\ref{fig:toy_memory_time}). Figure~\ref{fig:toy_mbar} instead holds $\batch$ fixed and shows how the reconstruction fills in with more iterations (larger $K$). We also track convergence and free energy estimates across more rarity levels $\eta \in \{5,6\}$ (Figure~\ref{fig:toy_more_etas}). More details in Appendix~\ref{app:toy_experiment}.

\begin{figure}[bp]
    \centering
    \begin{subfigure}[b]{0.17\linewidth}
        \includegraphics[width=\linewidth]{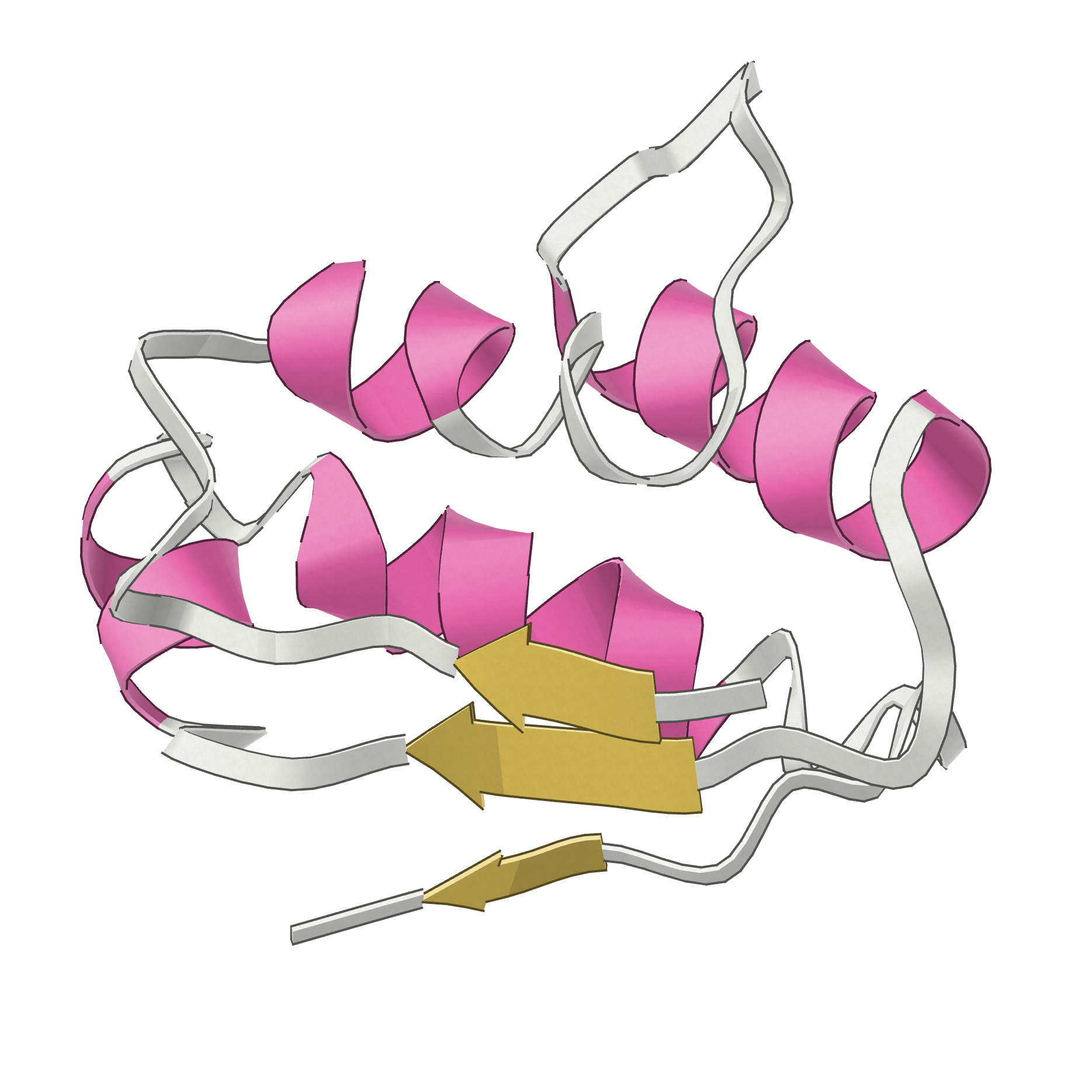}
        \caption{}\label{fig:struct_folded}
    \end{subfigure}\hfill
    \begin{subfigure}[b]{0.17\linewidth}
        \includegraphics[width=\linewidth]{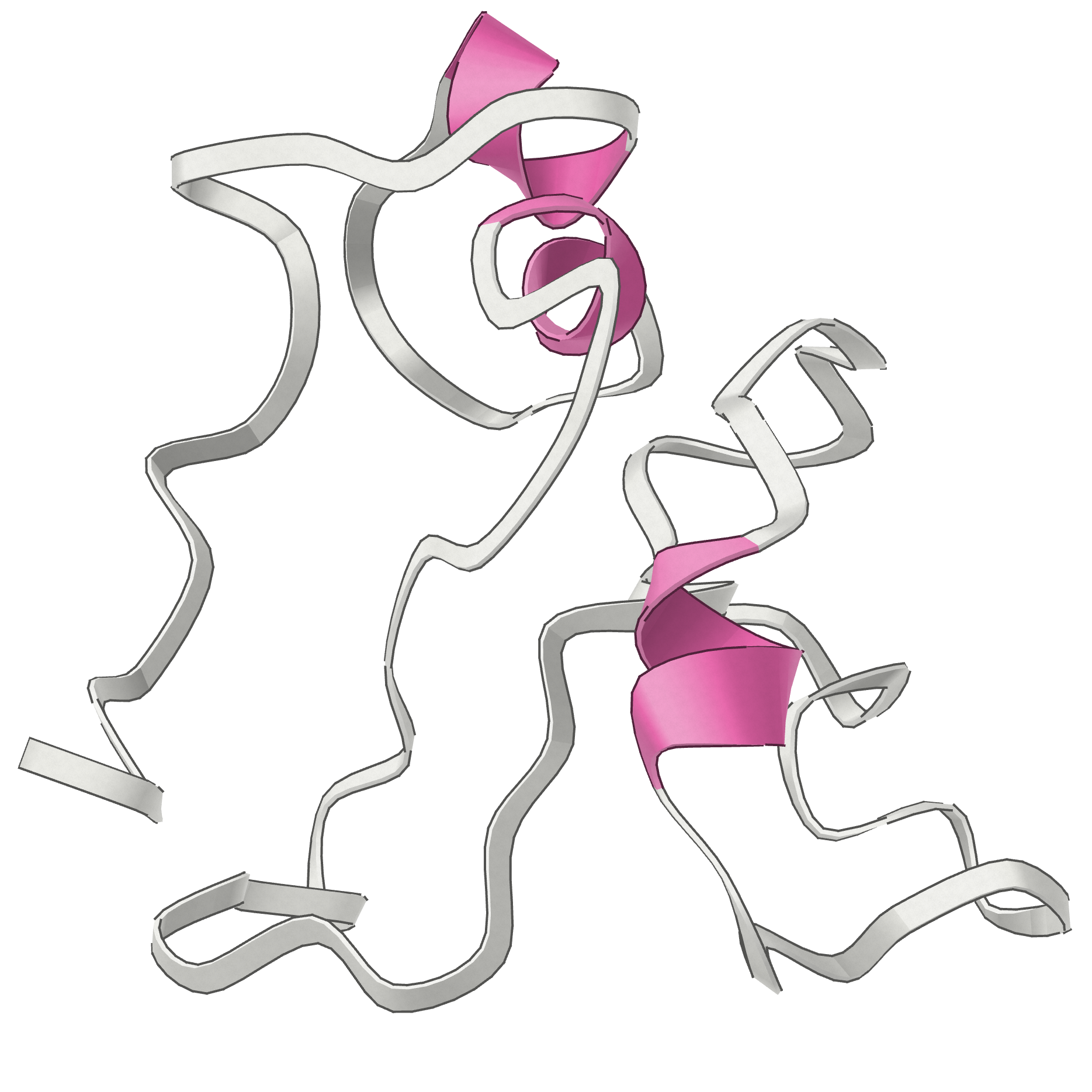}
        \caption{}\label{fig:struct_unfolded}
    \end{subfigure}\hfill
    \begin{subfigure}[b]{0.62\linewidth}
        \includegraphics[width=\linewidth]{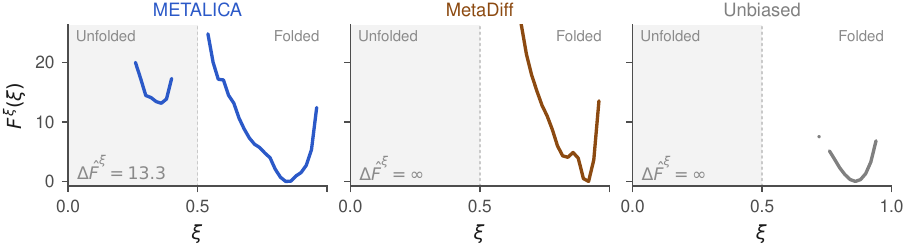}
        \caption{}\label{fig:protein_fnc}
    \end{subfigure}
    \caption{Free energy along the Fraction of Native Contacts for 1BTA. \textbf{(\subref{fig:struct_folded})} the folded state and \textbf{(\subref{fig:struct_unfolded})} an unfolded conformation generated by \metalica{}, colored by secondary structure: helices in pink, $\beta$-strands in yellow, loops and coil in white. \textbf{(\subref{fig:protein_fnc})} the free energy surface each method reconstructs; the shaded band is the unfolded state, $\cv < 0.5$.}
    \label{fig:protein}\figsq
\end{figure}

\parsq
\paragraph{Proteins.}
We further apply our enhanced sampling to BioEmu \citep{lewisScalableEmulationProtein2025}, a real-world diffusion model trained on a large collection of protein structures and fine-tuned on MD trajectories, capable of rapidly generating approximate conformational ensembles. However, for rare events such as protein unfolding, it still requires a dramatic number of samples to observe even a single unfolded conformation. Therefore, we leverage Fraction of Native Contacts (FNC) as our CV $\cv$, whose values $\cv(x)$ lie in $[0,1]$, to separate the folded state $\basinA = \{\cv > 0.5\}$ from the unfolded state $\basinB = \{\cv < 0.5\}$. We take 1BTA from the Protein Data Bank (PDB, \citet{bermanProteinDataBank2000}), among the most stable proteins in ProThermDB \citep{nikamProThermDBThermodynamicDatabase2021}. A separate pool of $20{,}000$ unbiased BioEmu draws for this protein contains a single unfolded structure. On the other hand, drawing $6{,}000$ samples using \metalica{} ($\batch = 150$) not only generates many unfolded structures, but also reveals a double-basin free energy landscape and estimates $\dfeh = 13.3$ with ${\approx}680$ rare samples generated (Figure~\ref{fig:protein}). More details about this experiment are in Appendix~\ref{app:protein_experiment}.

\secsq
\section{Discussion}\label{sec:discussion}
In this work, we showed how the Replica Exchange paradigm can diversely sample from a broad Collective Variable range through communications between replicas of a Markov chain ladder. Replacing batch parallelism with replica parallelism decouples accuracy from batch size: \metalica{}'s long sequential runs make free energy estimation affordable under memory constraints, and online refinement lets one chain follow the bias potential as it grows. \metalica{} does inherit the limitations of Metadynamics: it needs Collective Variables that separate the states of interest, and its samples are consecutive states of one chain rather than independent draws. Separately, what it estimates is the free energy landscape of the model that generated the samples, not of the underlying physical system.

\begin{ack}
JH acknowledges support from the University of Cambridge Harding Distinguished Postgraduate Scholars Programme. SS acknowledges the support of a CANSSI CRT grant and the NSERC Discovery Grant. JG acknowledges support from an NSERC grant. AO acknowledges support from the University of British Columbia Bioinformatics Graduate Program.
\end{ack}

\bibliographystyle{plainnat} %
\bibliography{references}

\clearpage
\appendix

\section{Well-Tempered Metadynamics}
\label{app:wtmetad}

Classical WTMetaD \citep{barducciWellTemperedMetadynamicsSmoothly2008} runs a
trajectory under the biased density \eqref{eq:metad} and, after each
sample $x_k$, deposits a single Gaussian hill of height $\hillh$ and width $\hillw$
centered at $\cv(x_k)$, damped by a bias factor $\biasfac > 1$ where the bias has
already grown:
\begin{equation}\label{eq:wt_deposit}
\hill_k(\cvvar) \;=\; \hillh\,
\exp\!\left(-\frac{\bias_{k}(\cv(x_k))}{\biasfac-1}\right)
\exp\!\left(-\frac{\norm{\cvvar - \cv(x_k)}^2}{2\hillw^2}\right),
\qquad
\bias_{k+1} = \bias_{k} + \hill_k .
\end{equation}
The first exponential is the tempering: hills shrink where the bias has already
accumulated, so $\bias_k$ converges. In the limit of many small hills the expected
increment at $\cvvar$ is proportional to the biased marginal times
$\exp(-\bias_k(\cvvar)/(\biasfac-1))$, whose fixed point satisfies
\begin{equation}\label{eq:wt_limit}
\fe(\cvvar) \;=\; -\frac{\biasfac}{\biasfac-1}\,\bias_\infty(\cvvar) + C,
\end{equation}
so the converged bias is a scaled negative of the free energy
\citep{barducciWellTemperedMetadynamicsSmoothly2008}. Standard Metadynamics
\citep{laioEscapingFreeenergyMinima2002} is $\biasfac \to \infty$.

\paragraph{Block deposition.}
Both methods emit $\batch$ samples per iteration and lay them down one hill at a time, each
hill tempered against the bias in force when it is laid rather than against the frozen
$\bias_k$. \metalica{}'s samples $x^{(n)}_k$ are unweighted states of one chain, so the
iteration's height budget $\hillh$ is split evenly across them: starting from
$\hill_k \equiv 0$ and applying, in order for $n = 1,\dots,\batch$,
\begin{equation}\label{eq:b_k}
\hill_k(\cvvar) \;\gets\; \hill_k(\cvvar) + \frac{\hillh}{\batch}
\exp\!\left(-\frac{\bias_{k}(\cv(x^{(n)}_k)) + \hill_k(\cv(x^{(n)}_k))}{\biasfac-1}\right)
\exp\!\left(-\frac{\norm{\cvvar - \cv(x^{(n)}_k)}^2}{2\hillw^2}\right),
\end{equation}
so each hill is damped by $\bias_k + \hill_k$, the bias already accumulated within the
block, and the iteration closes with $\bias_{k+1} = \bias_{k} + \hill_k$. Setting $\batch = 1$ recovers \eqref{eq:wt_deposit}. \metadiff{}
\citep{xieEnhancedDiffusionSampling2026} runs the same rule but splits $\hillh$ by its
particles' terminal importance weights rather than evenly. Damping by $\bias_k$ alone would
leave the fixed point \eqref{eq:wt_limit} unchanged but let a block place up to $\batch$
untempered hills in the same region, so the hills are taken in order to keep the deposition
rate well-tempered within an iteration as well as across iterations. \metalica{} uses
\eqref{eq:wt_limit} only as a heuristic for choosing $\hillh$ and $\biasfac$; the estimate
of $\fe$ comes from MBAR (Appendix~\ref{app:mbar}), which does not require the bias to have
converged.

\section{Sequential and parallel control of tilted diffusion paths}
\label{app:crepe}

This appendix uses the notation of the background: a tilted path
$\tpit{t} \propto \pit{t}\exp(\tbias{t})$ on a schedule $\schedule$, with $\tbias{t}$ from
\eqref{eq:tilted_path}. Iteration $k$ of \metalica{} is the instance $\tbias{0} = -\bias_k\circ\cv$, so that
$\tpit{t} = \pik{t}{k}$.

\paragraph{Proposals.}
Between levels $m-1$ and $m$ we use the forward noising process of the pretrained
model as $\fwd{m}{m-1}$ and, as $\bwd{m-1}{m}$, the reverse SDE integrated from $t_m$
to $t_{m-1}$ with the tilted score $\nabla_x\log\pit{t} + \nabla_x\tbias{t}$ of
\eqref{eq:tilted_path}; taking $\tbias{t} \equiv 0$ in the latter gives the pretrained
reverse kernel $\bwd{m-1}{m}^{0}$. Under the Euler-Maruyama discretization both
backward kernels are Gaussian transition densities, so their ratio is
available in closed form. The experiments run the tilted proposal throughout.

\paragraph{Computing the incremental weight.}
Substituting \eqref{eq:tilted_path} into \eqref{eq:incw},
\begin{equation}\label{eq:incw_full}
\begin{aligned}
\incw{m-1}{m}(x_{m-1}, x_m) \;=\;&
\exp\bigl(\tbias{t_{m-1}}(x_{m-1}) - \tbias{t_m}(x_m)\bigr)\\
&\times\;
\frac{\pit{t_{m-1}}(x_{m-1})\,\fwd{m}{m-1}(x_m \mid x_{m-1})}
     {\pit{t_m}(x_m)\,\bwd{m-1}{m}^{0}(x_{m-1} \mid x_m)}
\;\times\;
\frac{\bwd{m-1}{m}^{0}(x_{m-1} \mid x_m)}{\bwd{m-1}{m}(x_{m-1} \mid x_m)},
\end{aligned}
\end{equation}
up to the ratio of normalizing constants of $\tpit{t_{m-1}}$ and $\tpit{t_m}$, which
is common to all particles and cancels in normalized weights and in the swap ratio.
The middle factor is the ratio of the pretrained forward joint to the pretrained
backward joint. In continuous time the forward noising process and the reverse SDE
driven by $\nabla_x\log\pit{t}$ are time reversals of each other, so
$\pit{t}(x_t)\fwd{t'}{t}(x_{t'}|x_t) = \pit{t'}(x_{t'})\bwd{t}{t'}^{0}(x_t|x_{t'})$ and the middle factor
is one. With a discretized reverse SDE it is the Radon-Nikodym derivative between the
discrete forward and backward path measures, which \citet{heRNEPlugandplayDiffusion2025}
show can be evaluated along the move from the score alone (the RNE), and which
\crepe{} uses in its acceptance ratios \citep{heCREPECONTROLLINGDIFFUSION2026}. Because the proposal
carries the tilted score, the last factor does not vanish; it is evaluated in closed
form along the two simulated paths alongside the change in bias.

\paragraph{Sequential control.}
An SMC sampler with targets $\tpit{t_M},\dots,\tpit{t_0}$, proposals $\bwd{m-1}{m}$
and auxiliary backward kernels $\fwd{m}{m-1}$ has exactly \eqref{eq:incw} as its
incremental weight \citep{wuPracticalAsymptoticallyExact2023}: a particle at $x_m$ with
weight $\wt$ moves to $x_{m-1} \sim \bwd{m-1}{m}(\cdot\mid x_m)$ and takes weight
$\wt\,\incw{m-1}{m}(x_{m-1}, x_m)$. FKC \citep{skretaFeynmanKacCorrectorsDiffusion2025a} is the
continuous-time form of the same weight, and \metadiff{} uses it with the fixed control
drift $\tfrac12\sigma_t^2\nabla_x\tbias{t}$, which it obtains in the framework of
\citet{renDRIFTLITELIGHTWEIGHTDRIFT2026}.

\paragraph{Parallel control.}
The chain targets $\tpit{t_0}\times\cdots\times\tpit{t_M}$. For levels $(m-1,m)$
holding $(x^{m-1}, x^{m})$, the communication move proposes
$x'^{\,m} \sim \fwd{m}{m-1}(\cdot\mid x^{m-1})$ and
$x'^{\,m-1} \sim \bwd{m-1}{m}(\cdot\mid x^{m})$. This is a Metropolis-Hastings
proposal on the pair with kernel
$\fwd{m}{m-1}(x'^{\,m}\mid x^{m-1})\,\bwd{m-1}{m}(x'^{\,m-1}\mid x^{m})$,
so the acceptance ratio is
\begin{equation}\label{eq:swap_ratio_full}
\begin{aligned}
&\frac{\tpit{t_{m-1}}(x'^{\,m-1})\,\tpit{t_m}(x'^{\,m})}
     {\tpit{t_{m-1}}(x^{m-1})\,\tpit{t_m}(x^{m})}
\cdot
\frac{\fwd{m}{m-1}(x^{m}\mid x'^{\,m-1})\,\bwd{m-1}{m}(x^{m-1}\mid x'^{\,m})}
     {\fwd{m}{m-1}(x'^{\,m}\mid x^{m-1})\,\bwd{m-1}{m}(x'^{\,m-1}\mid x^{m})} \\
&\qquad=\;
\frac{\tpit{t_{m-1}}(x'^{\,m-1})\,\fwd{m}{m-1}(x^{m}\mid x'^{\,m-1})}
     {\tpit{t_m}(x^{m})\,\bwd{m-1}{m}(x'^{\,m-1}\mid x^{m})}
\cdot
\frac{\tpit{t_m}(x'^{\,m})\,\bwd{m-1}{m}(x^{m-1}\mid x'^{\,m})}
     {\tpit{t_{m-1}}(x^{m-1})\,\fwd{m}{m-1}(x'^{\,m}\mid x^{m-1})} ,
\end{aligned}
\end{equation}
where the right-hand side only regroups factors. The first fraction is
$\incw{m-1}{m}(x'^{\,m-1}, x^{m})$ by \eqref{eq:incw} and the second is
$\incw{m-1}{m}(x^{m-1}, x'^{\,m})^{-1}$, which is \eqref{eq:swap}. The identity holds for
any pair of proposal kernels. Sequential and parallel control therefore need exactly
the same ingredients: one backward simulation between adjacent levels and the
incremental weight of that move.

\paragraph{Schedule and boundary levels.}
The $M+1$ times of the schedule are the discretization grid of the reverse process
itself, uniformly spaced on $[0,1]$, so a level-to-level move is one step of the
integrator. Following \citet{syedNonReversibleParallelTempering2022}, step $n$
proposes exchanges on all pairs $(m-1,m)$ with $m \equiv n \pmod 2$; the proposals
within a step are independent and run in parallel. The top level has
$\tbias{1} \equiv 0$, so $\tpit{1} = \pit{1}$ and $\rep{M}{n}$ is refreshed by an exact
draw from $\pit{1}$ at every step, which is the source of fresh randomness carried
down the schedule by accepted exchanges. No local exploration move is applied at
intermediate levels in our experiments. The bottom particle $\rep{0}{n}$ is recorded
at every step as a sample from $\tpit{0}$.

\paragraph{Cost.}
Evaluating $\tbias{t_{m-1}}(x'^{\,m-1})$ requires $\denoiseat{t_{m-1}}(x'^{\,m-1})$, a
by-product of the score evaluation at the last sub-step of the backward proposal
that produced $x'^{\,m-1}$; evaluating $\tbias{t_m}(x'^{\,m})$ requires one further
score evaluation at the noised proposal. Per step the chain uses on the order of $M$
score evaluations, one per level traversed, matching the per-particle cost of one
SMC sweep over $M$ denoising steps. Since the chain is not reset when the bias
changes, the steps of iteration $k$ are productive almost from the start: the change
from $\pik{t}{k}$ to $\pik{t}{k+1}$ is the single increment in \eqref{eq:recursion},
whose scale shrinks under tempering, so the chain remains close to equilibrium for
the new target. A sequential sampler that reinitializes from noise at every $k$
instead spends $M$ denoising steps per particle before any sample from $\pik{0}{k}$
is available.

\section{Path recursion}
\label{app:recursion}

For $k = 0,\dots,K-1$, substituting $\bias_{k+1} = \bias_{k} + \hill_{k}$ into
\eqref{eq:intermediate_bias} gives
\begin{align}
\pik{t}{k+1}(x)
&\propto \pit{t}(x)\,\exp\Bigl(-\anneal_t\,\bias_{k+1}\bigl(\cv(\denoise(x))\bigr)\Bigr) \nonumber\\
&= \pit{t}(x)\,\exp\Bigl(-\anneal_t\,\bias_{k}\bigl(\cv(\denoise(x))\bigr)\Bigr)\,
   \exp\Bigl(-\anneal_t\,\hill_{k}\bigl(\cv(\denoise(x))\bigr)\Bigr) \nonumber\\
&\propto \pik{t}{k}(x)\,
   \exp\Bigl(-\anneal_t\,\hill_{k}\bigl(\cv(\denoise(x))\bigr)\Bigr),
\end{align}
which is \eqref{eq:recursion}.

\section{MBAR reweighting}
\label{app:mbar}

The pooled samples $\allsamples$ come from the $K+1$ thermodynamic states with reduced
potentials $u_k(x) = \pot(x) + \bias_{k}(\cv(x))$, $k = 0,\dots,K$, where
$\pot = -\log\pit{0}$ is the reduced potential of the unbiased model, and each state
contributes $\batch_k = \batch$ samples. \metadiff{}'s batch carries importance weights at
$m = 0$, so it is resampled to equal weights before pooling and the sums below are
unweighted for both methods. MBAR \citep{shirtsStatisticallyOptimalAnalysis2008a} solves the self-consistent equations for the
dimensionless free energies $\hat f_k$ of the states,
\begin{equation}\label{eq:mbar_f}
\exp(-\hat f_k) \;=\; \sum_{x \in \allsamples}
\frac{\exp\bigl(-\bias_{k}(\cv(x))\bigr)}
     {\sum_{\ell=0}^{K} \batch_\ell \exp\bigl(\hat f_\ell - \bias_{\ell}(\cv(x))\bigr)},
\end{equation}
where $\pot(x)$ has canceled between numerator and denominator, so the diffusion
model's density is never needed. The weight of sample $x$ under the unbiased target
$\pit{0}$ is then
\begin{equation}\label{eq:mbar_w}
\wt(x) \;\propto\;
\frac{1}{\sum_{\ell=0}^{K} \batch_\ell \exp\bigl(\hat f_\ell - \bias_{\ell}(\cv(x))\bigr)},
\end{equation}
normalized to sum to one, and the expectation of an observable $o$ under $\pit{0}$
is estimated by $\sum_x \wt(x)\,o(x)$. The CV marginal \eqref{eq:pmf} is the weighted histogram of
$\cv(x)$ and $\dfe$ is the log ratio of the total weight in $\basinA$ to that in
$\basinB$.

\section{\metalica{} implementation details}
\label{app:method}

\paragraph{Guidance schedule.}
The schedule $\anneal_t$ in \eqref{eq:tilted_path} shapes only the intermediate
targets and leaves $\pik{0}{k}$ unchanged, so it affects how well the chain mixes
across levels rather than what is estimated. We take it from the power family
\begin{equation}\label{eq:lambda}
\anneal_t \;=\; (1-t)^{\gexp}, \qquad \gexp > 0,
\end{equation}
where $\gexp = 1$ is linear interpolation and larger $\gexp$ concentrates the bias
into the levels closest to the
data. The values used are given with the rest of the settings in
Appendix~\ref{app:toy_model} and Appendix~\ref{app:protein_model}.

\paragraph{Burn-in.}
The chain is initialized once, by a single reverse pass from $\pit{1}$, and is cold
only at $k = 0$, where the target is the unmodified path $\pit{t}$. At the start of
every later iteration the state is kept and only the target changes, by the single
increment $\hill_k$ of \eqref{eq:recursion}.

\paragraph{Communication step in the code.}
The exchange probability \eqref{eq:swap} is computed from \eqref{eq:incw_full} along the
two simulated paths. Levels are the discretization grid of the reverse process itself, so
one exchange is one integrator step of size $\Delta t = t_m - t_{m-1}$, taken by
Euler-Maruyama on the positions, which follow the cosine VP-SDE with diffusion coefficient
$\beta_t = \sigma_t^2$ of \eqref{eq:sdes}. Writing $\beta_{\mathrm{hi}} = \beta_{t_m}$ and $\beta_{\mathrm{lo}} =
\beta_{t_{m-1}}$, the upward leg proposes $y \sim \mathcal{N}(x - \tfrac12
\beta_{\mathrm{lo}} x \Delta t,\; \beta_{\mathrm{lo}} \Delta t)$ and the downward leg proposes
from the tilted backward kernel, whose mean displaces the pretrained one by
$\beta_{\mathrm{hi}} \nabla_x \tbias{t}\,\Delta t$:
\begin{equation}\label{eq:kernels}
\mu^{\mathrm{P}} \;=\; y - \bigl(-\tfrac12 \beta_{\mathrm{hi}} y - \beta_{\mathrm{hi}}
\nabla_x\log\pit{t_m}(y)\bigr)\Delta t,
\qquad
\mu^{\mathrm{Q}} \;=\; \mu^{\mathrm{P}} + \beta_{\mathrm{hi}}\,\nabla_x \tbias{t_m}(y)\,\Delta t .
\end{equation}
The pretrained and tilted backward kernels share the same drift $\mu^{\mathrm{P}}$ and the
same variance $v = \beta_{\mathrm{hi}}\Delta t$, differing only by the displacement
$v\,\nabla_x\tbias{t_m}(y)$ in the mean, so the last factor of \eqref{eq:incw_full}
reduces to a difference of two Gaussian log-densities. That
difference is available in closed form,
\begin{equation}\label{eq:girsanov}
\log \bwd{m-1}{m}^{0}(x \mid y) - \log \bwd{m-1}{m}(x \mid y)
\;=\; -\,\bigl(x - \mu^{\mathrm{P}}\bigr)\cdot \nabla_x \tbias{t_m}(y)
\;+\; \frac{v}{2}\,\norm{\nabla_x \tbias{t_m}(y)}^2 ,
\end{equation}
so the acceptance needs no path integral beyond the scores already evaluated to make the
proposals. One detail matters for reproducing it. Each kernel's variance is taken at its
own conditioning time, $\beta_{\mathrm{hi}}\Delta t$ backward against
$\beta_{\mathrm{lo}}\Delta t$ forward, and the ratio is not invariant to writing a single
$\sigma_t^2$ for both.

\section{Experimental Details}
In this appendix, we provide technical details and additional results for the experiments we ran.

\subsection{Toy diffusion model}\label{app:toy_model}
The toy $\pit{0}$ is a Gaussian Mixture of the form
\begin{equation}
\pit{0}(x) = \sum_{i=1}^2 \frac{c_i}{\sigma_{\mathrm{mix}} \sqrt{2\pi}}
\exp\Biggl(-\frac{\left(x-\mu_i\right)^2}{2\sigma_{\mathrm{mix}}^2}\Biggr)
\end{equation}
with $\mu_1=-3$, $\mu_2=3$, $\sigma_{\mathrm{mix}}=0.5$, and mixture weights $c_1/c_2 = 10^\eta$. The second mode is the rare mode.

Since $\pit{0}$ is a Gaussian mixture, the base diffusion path $\{\pit{t}\}_{t\in[0,1]}$ can be constructed analytically. Under the variance-preserving forward map $x_t = a_t x_0 + \nu_t \epsilon$ with $\epsilon \sim \mathcal{N}(0,1)$, where $a_t$ is the signal coefficient and $\nu_t$ the noise level (both distinct from the guidance schedule $\anneal_t$ of \eqref{eq:lambda}), the marginal is also a two-component Gaussian mixture with the same weights,
\begin{align}
\pit{t}(x) &= \int \pit{0}(x_0)\,\mathcal{N}\left(x;a_t x_0, \nu_t^2\right)\dd x_0 \\
      &= \sum_{i=1}^{2} \frac{c_i}{s_t\sqrt{2\pi}}
         \exp\Biggl(-\frac{\left(x - a_t\mu_i\right)^2}{2 s_t^2}\Biggr),
\qquad s_t^2 = a_t^2\sigma_{\mathrm{mix}}^2 + \nu_t^2, \nonumber
\end{align}
so the means contract towards the origin as $a_t \to 0$ while the widths grow towards the prior. Writing the posterior responsibilities as
\begin{align}
q_i^t(x) = \frac{c_i\,\mathcal{N}\left(x;a_t\mu_i, s_t^2\right)}
                {\sum_{j=1}^{2} c_j\,\mathcal{N}\left(x;a_t\mu_j, s_t^2\right)},
\end{align}
the score function is then available in closed form as a responsibility-weighted average of the two component scores,
\begin{align}
\nabla_x \log\pit{t}(x) = -\sum_{i=1}^{2} q_i^t(x)\,\frac{x - a_t\mu_i}{s_t^2},
\end{align}
which is used in place of a learned score network so that no network approximation error enters the comparison.

We used the Variance Preserving Stochastic Differential Equation (VP-SDE) formulation of diffusion models \citep{songSCOREBASEDGENERATIVEMODELING2021} with the cosine noise schedule, $a_t = \cos\!\bigl(\tfrac{\pi}{2}\tfrac{t+\delta}{1+\delta}\bigr)/ \cos\!\bigl(\tfrac{\pi}{2}\tfrac{\delta}{1+\delta}\bigr)$ with $\delta = 0.008$ and $\nu_t = \sqrt{1-a_t^2}$, which is the schedule the protein model also uses. For inference, we use the Euler-Maruyama algorithm \citep{songSCOREBASEDGENERATIVEMODELING2021} with $M = 100$ discrete steps. We set the bias factor $\biasfac = 10$, the hill width $\hillw = 0.5$, and the guidance exponent $\gexp = 0.5$. The CV grid spans $\cvvar \in [-8,8]$ on $321$ points; the figures are drawn over $[-6,6]$.

\subsection{Protein diffusion model}\label{app:protein_model}
We used BioEmu \citep{lewisScalableEmulationProtein2025} (checkpoint \texttt{bioemu-v1.1}), a diffusion model conditioned on a single amino-acid sequence through an Evoformer embedding and trained on protein structures, then fine-tuned on MD trajectories. BioEmu represents a protein as a sequence of residue frames, a backbone $\mathrm{C}\alpha$ position and an orientation per residue, noised by separate processes; the positions follow the same cosine VP-SDE as the toy. For inference, BioEmu uses a second-order DPM-Solver++ over $M = 100$ steps on $t \in [10^{-3}, 0.99]$, taking a midpoint prediction and re-evaluating the score there.

We implemented \metadiff{} on top of this solver by attaching the FKC correction to the same $M$ steps: at each level $t$ the proposal drift carries the reward at half the weight of the score, $\nabla_x \log \pit{t}(x) + \tfrac12 \nabla_x \tbias{t}(x)$ with $\tbias{t}(x) = -\anneal_t \bias_k\bigl(\cv(\denoise(x))\bigr)$ as in \eqref{eq:intermediate_bias}, which is the corrector's own equation rather than the exact tilted reverse SDE, evaluated on the CV of the predicted clean structure, and the log-weight of each particle is incremented by the corresponding Feynman-Kac term, with the population resampled whenever its $\ESS$ falls below $0.5$ of the batch \citep{xieEnhancedDiffusionSampling2026}.

We implemented \metalica{} by placing one replica on each of the $M+1 = 101$ solver times and integrating the exchange proposal by Euler-Maruyama. Positions follow the cosine VP-SDE of the toy, so their forward and backward kernels are Gaussian and their densities are available in closed form. Both methods use the tilt exponent $\gexp = 5$.

We use $\biasfac = 10$, an initial height $\hillh = 1$, a hill width $\hillw = 0.03$ in FNC units and a CV grid of $200$ points, and draw $K = 40$ iterations of $\batch = 150$ samples, from the same seed in both methods, \metadiff{} and \metalica{}.

\subsection{Additional results for the toy experiments}\label{app:toy_experiment}

\begin{figure}[htbp]
    \centering
    \includegraphics[width=0.85\linewidth]{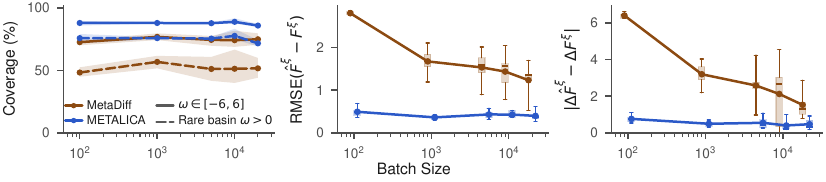}
    \caption{Reconstruction quality against the iteration sample size $\batch$ at fixed sample budget. \textbf{Left:} fraction of the CV grid resolved, over the whole window (solid) and the rare basin (dashed). \textbf{Center:} RMSE of $\feh - \fe$. \textbf{Right:} error in $\dfeh$. Boxes are seeds, joined markers their means.}
    \label{fig:toy_bs}
\end{figure}

\begin{figure}[htbp]
    \centering
    \includegraphics[width=0.6\linewidth]{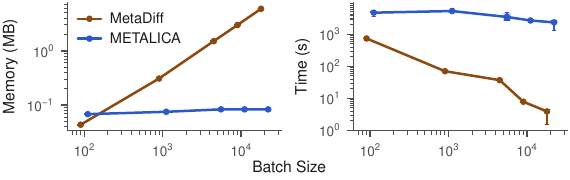}
    \caption{Cost against the iteration sample size $\batch$ at fixed sample budget: peak GPU memory (left) and sampling time (right), both logarithmic. Boxes are seeds, joined markers their means.}
    \label{fig:toy_memory_time}
\end{figure}

\begin{figure}[htbp]
    \centering
    \includegraphics[width=\linewidth]{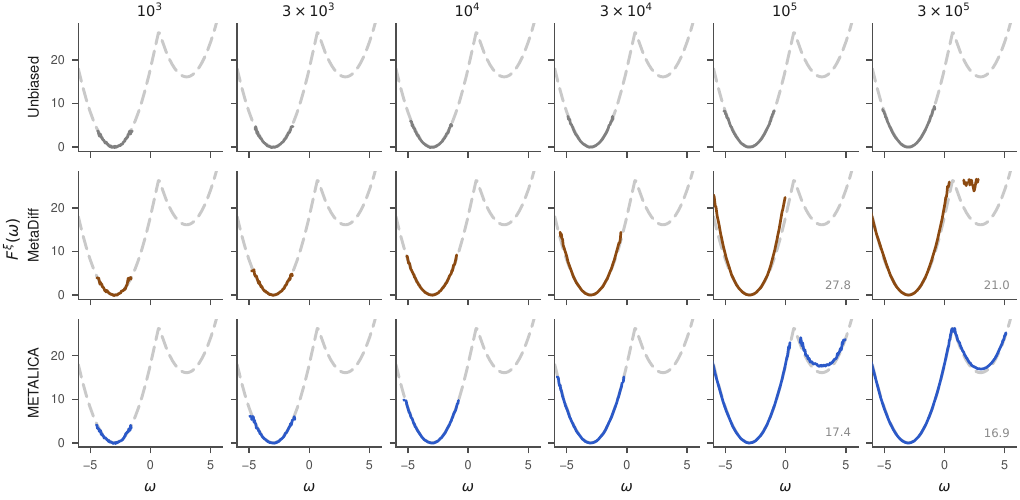}
    \caption{Reconstructed surfaces at $\dfe = 16.1$, one row per method and one column per cumulative sample count, against the true $\fe$ (dashed). Each cell is one run truncated to its first $n$ draws, at a fixed iteration sample size.}
    \label{fig:toy_mbar}
\end{figure}

\begin{figure}[htbp]
    \centering
    \includegraphics[width=\linewidth]{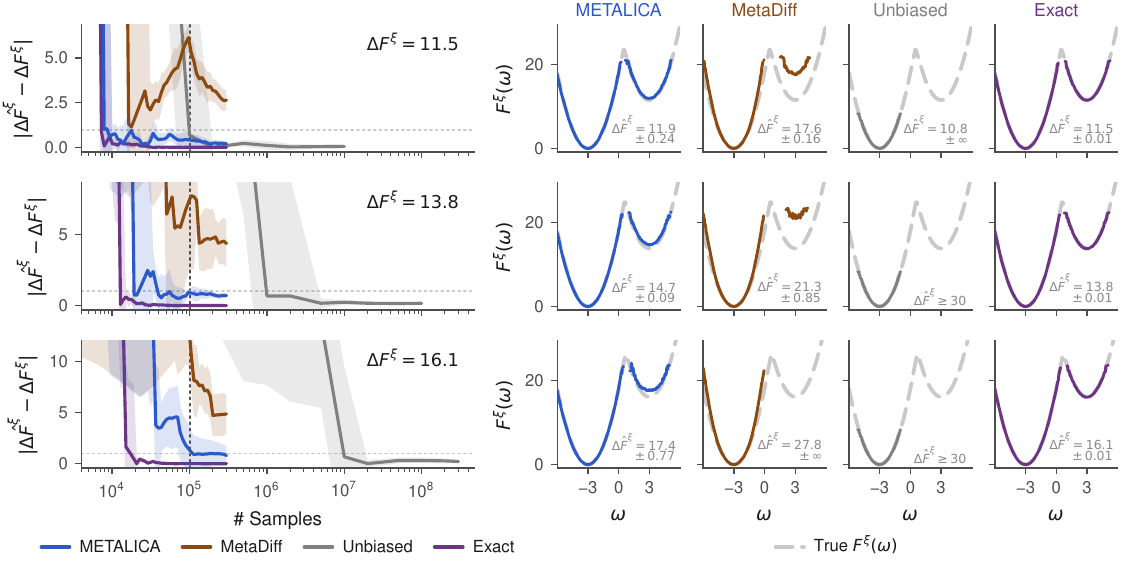}
    \caption{As Figure~\ref{fig:toy}, at three rarity levels (rows) and with the Exact control added as a fourth profile column. \textbf{Left:} error in $\dfeh$ against cumulative samples, median over seeds with band. \textbf{Right:} $\feh$ at the deposition budget against the true $\fe$ (dashed).}
    \label{fig:toy_more_etas}
\end{figure}

The final control is the exact \metalica{}/\metadiff{} version where the sampler is not a diffusion model, but an exact analytical sampler that uses the inverse-CDF method. This ``Exact'' version provides an upper bound on performance. Figures~\ref{fig:toy_bs}--\ref{fig:toy_more_etas} report the reconstruction sweep, its cost, the fill-in with more iterations, and the two additional rarity levels.

\subsection{Additional results for the protein experiment}\label{app:protein_experiment}
The \metalica{} run drew $6{,}000$ samples in $40$ iterations of $\batch = 150$ on one GPU. \metadiff{}, run on the same system, the same CV, the same seed and the same $40\times150$ budget, placed none of its $6{,}000$ samples below $\cvvar = 0.5$, so it has no rare basin to reconstruct and its $\dfeh$ is unbounded.

\end{document}